\documentclass[11pt,a4paper]{article}
\usepackage[hyperref]{emnlp2020}
\usepackage{times}
\usepackage{latexsym}

\usepackage{graphicx}
\usepackage{xspace}
\usepackage{algorithm}
\usepackage{algorithmic}
\usepackage{amsmath}
\usepackage{amssymb}
\usepackage{booktabs}
\usepackage{multirow}
\usepackage{subfigure}

\newcommand{\method}{AUSDS\xspace}
\newcommand{\framework}{\method learning framework\xspace}

\usepackage{microtype}

\aclfinalcopy 
\def\aclpaperid{3602} 

\title{Active Sentence Learning by Adversarial Uncertainty Sampling\\ in Discrete Space}

\author{
\parbox{\linewidth}{
\centering
Dongyu Ru$^{\dagger,\ddagger}$, Jiangtao Feng$^\dagger$, 
Lin Qiu$^\ddagger$, 
Hao Zhou$^\dagger$,\\
Mingxuan Wang$^\dagger$, 
Weinan Zhang$^\ddagger$, 
Yong Yu$^\ddagger$, 
Lei Li$^\dagger$
}\\
$^\dagger$ByteDance AI Lab \\
{\small\texttt{\{fengjiangtao,zhouhao.nlp,wangmingxuan.89,lileilab\}@bytedance.com}}\\
  $^\ddagger$Shanghai Jiao Tong University\\
  {\small\texttt{\{maxru,lqiu,wnzhang,yyu\}@apex.sjtu.edu.cn}} 
}

\date{}

\begin{document}
\maketitle
\begin{abstract}
Active learning for sentence understanding aims at discovering informative unlabeled data for annotation and therefore reducing the demand for labeled data. 
We argue that the typical uncertainty sampling method for active learning is time-consuming and can hardly work in real-time, which may lead to ineffective sample selection. 
We propose adversarial uncertainty sampling in discrete space (\method) to retrieve informative unlabeled samples more efficiently.
\method maps sentences into latent space generated by the popular pre-trained language models, and discover informative unlabeled text samples for annotation via adversarial attack. 
The proposed approach is extremely efficient compared with traditional uncertainty sampling with more than 10x speedup. 
Experimental results on five datasets show that \method outperforms strong baselines on effectiveness.
\end{abstract}

\section{Introduction}
\label{sec:intro}

Deep neural models become popular in natural language processing~\cite{peters2018deep, radford2018improving, devlin2018bert}.
Neural models usually consume massive labeled data, which requires a huge quantity of human labors.
But data are not born equal, where informative data with high uncertainty are decisive to decision boundary and are worth labeling.
Thus selecting such worth-labeling data from unlabeled text corpus for annotation is an effective way to reduce the human labors and to obtain informative data.

Active learning approaches are a straightforward choice to reduce such human labors.
Previous works, such as uncertainty sampling~\cite{lewis1994sequential}, needs to traverse all unlabeled data to find informative unlabeled samples, which are always near the decision boundary with large entropy.
However, the traverse process is very time-consuming, thus cannot be executed frequently~\cite{settles2008analysis}.
A common choice is to perform the sampling process after every specific period, and it samples and labels informative unlabeled data then trains the model until convergence~\cite{deng2018adversarial}.

\begin{figure*}[t]
\centering
\includegraphics[width=0.9\textwidth]{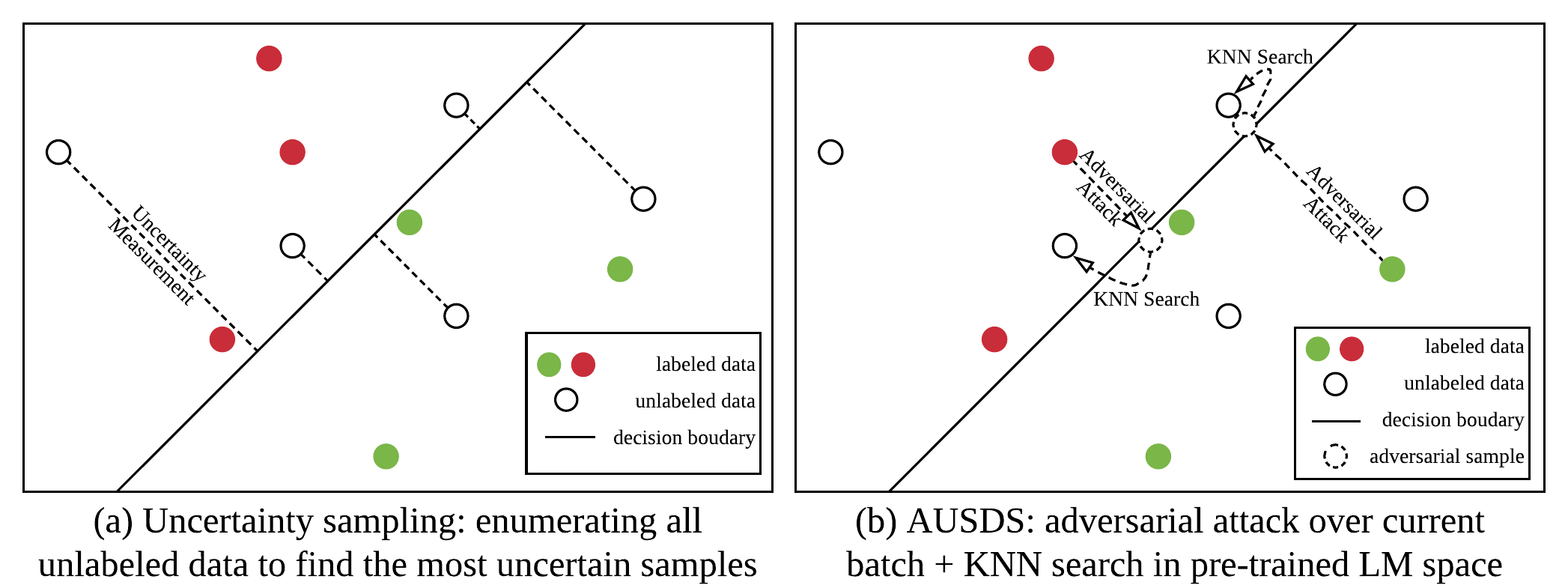}
\caption{Comparison between uncertainty sampling and \method for active learning.}
\label{fig:alvsaaal}
\end{figure*}

We argue that infrequently performing uncertainty sampling may lead to the ``ineffective sampling'' problem.
Because in the early phase of training, the decision boundary changes quickly, which makes previously collected samples less effective after several updates of the model.
Ideally, uncertainty sampling should be performed frequently in the early phase of model training.

In this paper, we propose the adversarial uncertainty sampling in discrete space~(\method) to address the ineffective sampling problem for active sentence learning by introducing more frequent sampling with significantly lower costs.
Specifically, we propose to leverage the adversarial attack~\cite{goodfellow2014explaining,kurakin2016adversarial} to the selecting of informative samples with high uncertainty, which significantly narrows down the search space.
Fig.~\ref{fig:alvsaaal} shows the difference between uncertainty sampling and \method.
The typical uncertainty sampling (Fig.~\ref{fig:alvsaaal}.a) traverses all the unlabeled samples to obtain samples of high uncertainty for each sampling run, which is costly with time complexity ($O({\rm Unlabeled~Data~Size})$.
\method (Fig.~\ref{fig:alvsaaal}.b) first projects a labeled text to the decision boundary, denoted as an adversarial data point, and searches nearest neighbors of this point.
The computational cost of \method is significantly smaller than typical uncertainty sampling with the time complexity $O({\rm Batch~Size})$.
But it is non-trivial for \method to perform adversarial attacks, which requires adversarial gradients on sentences, since texts live in a discrete space.
We propose to include a pre-trained neural encoder, such as BERT \cite{devlin2018bert}, to map unlabeled sentences into a continuous space, over which the adversarial attack is performed.
Since not every adversarial data point in the encoding space can be mapped back to one of the unlabeled sentences, we propose to use the k-nearest neighbor (KNN) algorithm~\cite{altman1992introduction} to find the most similar unlabeled sentences~(the adversarial samples) to the adversarial data points.
Besides, empirically, we mix some random samples into the uncertainty samples to alleviate the sampling bias issue mentioned by~\cite{huang2010active}.
Finally, the mixed samples are sent to an oracle annotator to obtain their label and are appended to the labeled data set.

We deploy \method for active sentence learning and conduct experiments on five datasets across two NLP tasks, namely sequence classification and sequence labeling. 
Experimental results show that \method outperforms random sampling and uncertainty sampling strategies.

Our contributions are summarized as follows:
\begin{itemize}
\setlength{\itemsep}{0pt}
    \item We propose \method for active sentence learning, which first introduces the adversarial attack for sentence uncertainty sampling, alleviating the ineffective sampling problem.
    \item We propose to map sentences into the pre-trained LM encoding space, which makes adversarial uncertainty sampling available in the discrete sentence space. 
    \item Experimental results demonstrate that our active sentence learning framework by \method, which we call \framework, outperforms strong baselines in sampling effectiveness with acceptable running time.
\end{itemize}

\section{Related Work}
\label{sec:related}
This work focuses on reducing the labeled data size with the help of pre-trained LM in solving sentence learning tasks. 
The proposed \method approach is related to two different research topics, active learning and adversarial attack.
\begin{figure*}[ht]
\centering
\includegraphics[width=0.9\textwidth]{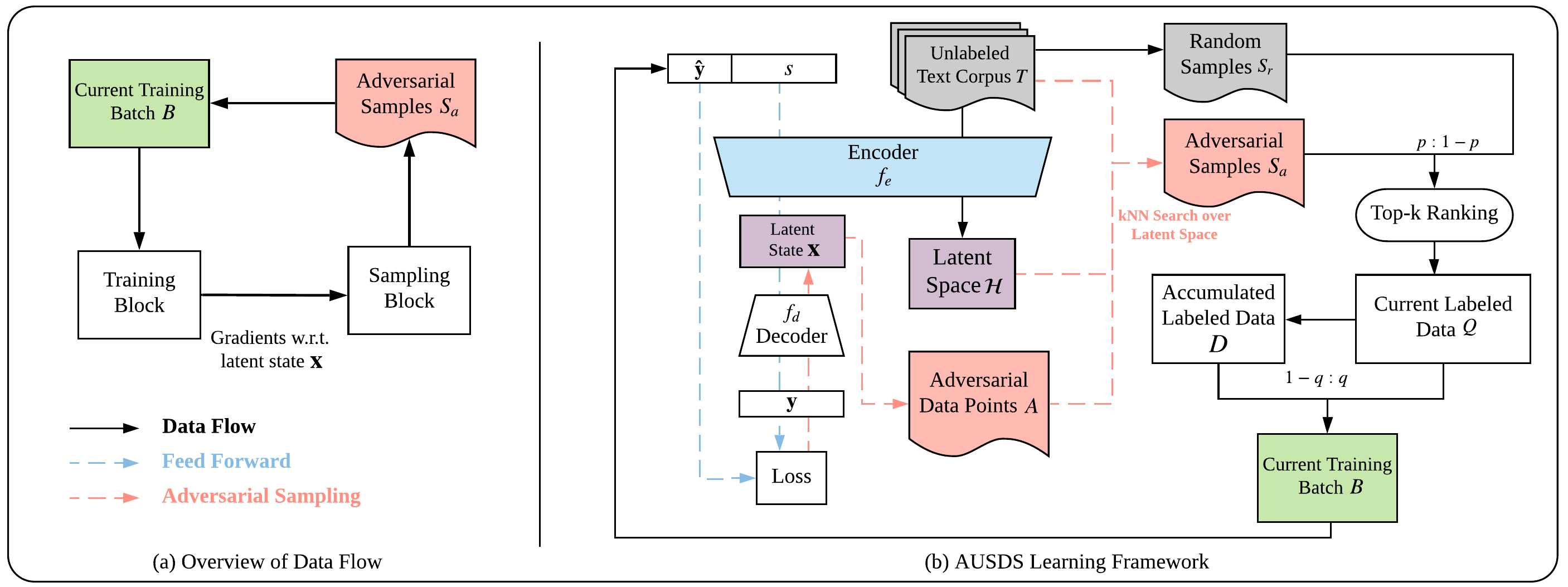}
\caption{Overview of active sentence learning framework by AUSDS. Some notations are labeled along with corresponding components.}
\label{fig:framework}
\end{figure*}

\subsection{Active Learning}
Active learning algorithms can be categorized into three scenarios, namely membership query synthesis, stream-based selective sampling, and pool-based active learning~\cite{settles2009active}.
Our work is more related to pool-based active learning, which assumes that there is a small set of labeled data and a large pool of unlabeled data available~\cite{lewis1994sequential}. 
To reduce the demand for more annotations, the learner starts from the labeled data and selects one or more queries from the unlabeled data pool for the annotation, then learns from the new labeled data and repeats.

The pool-based active learning scenario has been studied in many real-world applications, such as text classification~\cite{lewis1994sequential,hoi2006large}, information extraction~\cite{settles2008analysis} and image classification~\cite{joshi2009multi}.
Among the query strategies of existing active learning approaches, the uncertainty sampling strategy~\cite{joshi2009multi,lewis1994sequential} is the most popular and widely used.
The basic idea of uncertainty sampling is to enumerate the unlabeled samples and compute the uncertainty measurement like information entropy for each sample.
The enumeration and uncertainty computation makes the sampling process costly and cannot be performed frequently, which induced the ineffective sampling problem.

There are some works that focus on accelerating the costly uncertainty sampling process. 
\citet{jain2010hashing} propose a hashing method to accelerate the sampling process in sub-linear time. 
\citet{deng2018adversarial} propose to train an adversarial discriminator to select informative samples directly and avoid computing the rather costly sequence entropy.
Nevertheless, the above works are still computationally expensive and cannot be performed frequently, which means the ineffective sampling problem still exists.

\subsection{Adversarial Attack}
Adversarial attacks are originally designed to approximate the smallest perturbation for a given latent state to cross the decision boundary~\cite{goodfellow2014explaining,kurakin2016adversarial}.
As machine learning models are often vulnerable to adversarial samples, adversarial attacks have been used to serve as an important surrogate to evaluate the robustness of deep learning models before they are deployed~\cite{biggio2013evasion,szegedy2013intriguing}.
Existing adversarial attack approaches can be categorized into three groups, which are one-step gradient-based approaches~\cite{goodfellow2014explaining,rozsa2016adversarial}, iterative methods~\cite{kurakin2016adversarial} and optimization-based methods~\cite{szegedy2013intriguing}. 

Inspired by the similar goal of adversarial attacks and uncertainty sampling, in this paper, instead of considering adversarial attacks as a threat, we propose to combine these two approaches for achieving real-time uncertainty sampling.
Some works share a similar but different idea with us.
\citet{li2018query} introduce active learning strategies into black-box attacks to enhance query efficiency.
\citet{pal2020activethief} also use active learning strategies to reduce the number of queries for model extraction attacks.
\citet{zhu2017generative} propose to train Generative Adversarial Networks to generate samples by minimizing the distance to the decision boundary directly, which is in the query synthesis scenario different from us.
\citet{ducoffe2018adversarial} also introduce adversarial attacks into active learning by augmenting the training set with adversarial samples of unlabeled data, which is infeasible in discrete space.
Note that none of the works above share the same scenario with our problem setting. 

\section{Active Sentence Learning with \method}
\label{sec:method}
\begin{algorithm*}[ht]
\caption{Active Sentence Learning with Adversarial Uncertainty Sampling in Discrete Space}
\label{alg:aaal}
\textbf{Input:} an unlabeled text corpus $T_0$, an oracle $O$, a labeled data $D_0=\{(s, O(s))|s\in S_0, \text{ a small initial text corpus}\}$, pre-trained LM $f_e$, fine-tuning interval $j$, and fine-tuning step $k$.\\
\textbf{Output:} a well-trained model $f = (f_e, f_d)$
\begin{algorithmic}[1]
\STATE Train $f_d$ on $D_0$ with frozen $f_e$;
\STATE Construct a discrete bijective mapper $M$, where $M(s)=f_e(s) \in \mathcal{H}$ and $M^{-1}(f_e(s))=s\in T_0$;
\STATE Sample a training batch $B_0$ from $D_0$;
\STATE $i \leftarrow 0$
\WHILE{$|T_i| > 0$}
  \STATE Train decoder $f_d$ on $B_i$ with frozen encoder $f_e$;
  \STATE Generate adversarial data points $A \subset \mathcal{H}$ using the adversarial attack algorithm;
  \STATE Retrieve adversarial samples $S_a=\{s_a=M^{-1}(x) \in T_i | x \in \text{KNN}(A)\}$;
  \STATE Inject $S_a$ with random samples $S_r$, where $|S_a| : |S_r| = p : 1-p$;
  \STATE Select top-k ranked samples $S_{add}$ from $S_a$ w.r.t. the information entropy;
  \STATE Label new data $Q \leftarrow \{(s, O(s)) | s \in S_{add}\}$;
  \STATE Update labeled data $D_{i+1} \leftarrow D_i \cup Q$;
  \STATE Remove newly labeled data from unlabeled dataset $T_{i+1} \leftarrow T_i - S_{add}$;
  \STATE Sample a training batch $B_{i+1}$ from $Q$ and $D_{i+1}$ by the ratio of $q : 1-q$;
  \IF{$i$ mod $j = 0$}
    \STATE Fine-tune $f$ with $D_{i+1}$ for $k$ steps;
    \STATE Update the mapper $M$ with the fine-tuned encoder $f_e$ and text corpus $T_{i+1}$;
  \ENDIF
  \STATE $i \leftarrow i+1$
\ENDWHILE
\end{algorithmic}
\end{algorithm*}

We propose \framework, an efficient and effective computational framework for active sentence learning.
The overview of the learning framework is shown in Fig.~\ref{fig:framework}.
The learning framework consists of two blocks, a training block and a sampling block \method.
The training block learns knowledge from the labeled data, whereas the sampling block retrieves valuable unlabeled samples, whose latent states are close to the decision boundary over the latent space, from the unlabeled text corpus.
Note that the definition of latent spaces can be different across encoders and tasks.
The samples retrieved by the sampling block will be further sent to an oracle annotator to obtain their label, and the new samples with labels are also appended to the labeled data.

In this section, we first introduce \method method by showing how \method select samples that are critical to the decision boundary over the latent space.
Then we present the computational procedure of the full-fledged framework in detail.

\subsection{\method}
\label{sec:ausds}

\method first defines a latent space, over which sentences are distinguishable according to the model's decision boundary. 
The latent space is usually determined by the encoder architecture and the downstream task.
We detail the latent space definition of specific encoders and tasks in Sec.~\ref{sec:exp:set-up}.


At first, we sample a batch of labeled texts and compute their representation as well as their gradients in the latent space.
Using the latent states and their gradients, we perform adversarial attacks to generate adversarial data points $A$ near the decision boundary in the latent space.
Adversarial attacks are performed using the following existing approaches:
\begin{itemize}
    \item {Fast Gradient Value (FGV)~\cite{rozsa2016adversarial}: a one-step gradient-based approach with high efficiency. The adversarial data points are generated by:
    \begin{equation}
        \setlength{\abovedisplayskip}{2pt}
        \setlength{\belowdisplayskip}{2pt}
        \mathbf{x}' =  \mathbf{x} + \lambda \cdot \nabla_\mathbf{x} F_d(\mathbf{x})
        \label{eq:attack}
    \end{equation}
    where $\lambda$ is a hyper parameter, and $F_d$ is the cross entropy loss on $\mathbf{x}$.
    }
    \item {DeepFool \cite{Moosavi-Dezfooli_2016_CVPR}: an iterative approach to find the minimal perturbation that is sufficient to change the estimated label.}
    \item {C\&W \cite{carlini2017towards}: an optimization-based approach with the optimization problem defined as:
    \begin{equation}
        \setlength{\abovedisplayskip}{2pt}
        \setlength{\belowdisplayskip}{2pt}
        \text{minimize } D(\mathbf{x}, \mathbf{x}') + c\cdot g(\mathbf{x}')
    \end{equation}
    where $g(\cdot)$ is a manually designed function, satisfying $g(\mathbf{x}) \le 0$ if and only if $\mathbf{x}$'s label is a specific target label. $D$ is a distance measurement like Minkowski distance.
    }
\end{itemize}
FGV is efficient in the calculation, whereas the other two methods typically find more precise adversarial data points but with larger computational costs. 
We use all of them in our experimental part to show the effectiveness of the \method. 

In our sentence learning scenario, the adversarial data points $A$ cannot be grounded on real natural language text samples.
Thus we perform k-nearest neighbor (KNN) search~\cite{altman1992introduction} to find unlabeled text samples whose latent states are k-nearest to the adversarial data points $A$.

We implement the KNN search using Faiss\footnote{https://github.com/facebookresearch/faiss}~\cite{johnson2017billion}, an efficient similarity search algorithm with GPUs.
The computational cost of KNN search results from two processes, including constructing a sample mapper $M$ between text and latent space, and searching similar latent states of adversarial data points.
The sampler mapper $M$ here is constructed as a hash map, which is of high computational efficiency, to memorize the mapping between an unlabeled text $s$ and its latent representation $\mathbf{x}$.
The sample mapper is only reconstructed when the encoder is updated, and infrequent encoder updates contribute to efficiency.
Besides, the searching process is also fast (100$\times$ faster than generating $A$) thanks to Faiss.
Thus it is possible to performed \method frequently at batch-level without harming computation.

After acquiring adversarial samples $S_a$ using KNN search, we mix $S_a$ with random samples $S_r$ drawn from unlabeled text corpus $T_i$ by the ratio of $p : 1-p$, where $p$ is a hyper-parameter determined on the development set.
The motivation of appending random samples is to balance exploration and exploitation, thus avoiding the model continuously retrieve samples in a small neighborhood.

We perform top-k ranking over the information entropy of the mixed samples to further retrieve samples with higher uncertainty. 
Since the size of the mixed samples is comparable to the batch size, the computation cost is acceptable. 
The remaining samples are further sent to an oracle annotator $O$ to obtain their labels.

\subsection{Active Learning Framework}

The overall procedure of the proposed framework equipped with \method is outlined in Algorithm~\ref{alg:aaal}

\paragraph{Initialization}
The initialization stage is shown in Algorithm~\ref{alg:aaal} line 1-4. 
We first initialize our encoder $f_e$ with the pre-trained LM, which can be ${\rm BERT}_{\rm BASE}$~\cite{devlin2018bert} or ${\rm ELMo}$~\cite{peters2018deep}.
The decoder here is built upon the latent space and is randomly initialized.
After building up the neural model architecture, we train only the decoder on existing labeled data $D_0$ to compute an initial decision boundary on the latent space.
Meanwhile, we construct an initial discrete sample mapper $M$ used for the sampling block.
Finally, we sample a training batch $B_0$ from labeled data corpus $D_0$, and set current training step $i$ to $0$.

\paragraph{Training}
The training stage is shown in Algorithm~\ref{alg:aaal} line 6.
With the defined decoders $f_d$ and a training batch $B_{i}$, we train the decoder with a cross entropy loss (Fig.~\ref{fig:framework}.b).
Note that during the training process, we freeze the encoder as well as the latent space, where a frozen latent space contributes to computational efficiency without reconstructing the mapper $M$.

\paragraph{Sampling}
The sampling stage is shown in Algorithm~\ref{alg:aaal} line 7-14.
As is shown in Sec.~\ref{sec:ausds}, given the gradients on the current batch $B_i$ w.r.t. latent states during training, the sampling process generates the adversarial samples $S_a$ and labels the samples with high uncertainty from a mixture of $S_a$ and randomly injected unlabeled data $S_r$.
The labeled samples $Q$ are removed from the unlabeled text corpus and inserted into labeled data, resulting in $T_{i+1}$ and $D_{i+1}$ respectively.
Then we create a new training batch consist of samples from $Q$ and $D_{i+1}$ with a ratio of $q:1-q$, which favors the newly selected data $Q$, because the newly selected ones are considered as more critical to the current decision boundary.

\paragraph{Fine-Tuning}
The fine-tuning stage is shown in Algorithm~\ref{alg:aaal} line 15-18. 
We fine-tune the encoder for $k$ steps after $j$ batches are trained.
During the fine-tuning process, both of the encoder and the decoder are trained on the accumulated labeled data set $D_{i+1}$. The encoder is also fine-tuned for enhancing overall performance. Experiments show that the final performance is harmed a lot without updating the encoder.
Then we update the mapper $M$ for the future KNN search, because the fine-tuning of the encoder corrupts the projection from texts to latent spaces, which requires renewal of the sampler mapper $M$.
The algorithm terminates until the unlabeled text corpus $T_i$ is used up.


\section{Experiments}
\label{sec:exp}
We evaluate the \framework on sequence classification and sequence labeling tasks.
For the oracle labeler $O$, we directly use the labels provided by the datasets.
In all the experiments, we take average results of 5 runs with different random seeds to alleviate the influence of randomness.
\begin{table*}[t]\small
    \centering 
    \begin{tabular}{lcr}
        \toprule
        Dataset & Task & Sample Size\\ 
        \hline
        SST-2~\cite{socher2013recursive}& sequence classification & 11.8k sentences, 215k phrases\\
        SST-5~\cite{socher2013recursive}& sequence classification & 11.8k sentences, 215k phrases\\
        MRPC~\cite{dolan2004unsupervised}& sequence classification & 5,801 sentence pairs\\
        AG News~\cite{zhang2015character}& sequence classification & 12k sentences\\
        CoNLL'03~\cite{sang2003introduction} & sequence labeling & 22k sentences, 300k tokens\\
        \bottomrule
    \end{tabular}
    \caption{5 datasets we used for sentence learning experiments, across sequence classification and sequence labeling tasks.}
    \label{table:dataset}
\end{table*}
\begin{table*}[t]\small
    \centering
    \begin{tabular}{cccccc}
    \toprule
    Dataset & RM & US & AUSDS(FGV) & AUSDS(DeepFool) & AUSDS(C\&W) \\ \midrule
    SST-2     & 1061x & 1x & 38x & 38x & 28x \\
    SST-5     & 1939x & 1x & 52x & 52x & 38x \\
    MRPC      & 97x & 1x & 14x & 14x & 11x \\
    AG News   & 1434x & 1x & 51x & 47x & 38x\\
    CoNLL'03  & 45x & 1x & 10x & --- & --- \\ \bottomrule
    \end{tabular}
    \caption{The average speedup of each sampling step in comparison with US on 5 datasets with BERT as the encoder. The statistics are collected using Tesla-V100 GPU. US scans the unlabeled data once when 2\% of data are labeled. The AUSDS using DeepFool and C\&W on CoNLL'03 are omitted because these adversarial attack methods are not suitable for sequence labeling.}
    \label{tab:samp_speed}
\end{table*}

\subsection{Set-up}
\label{sec:exp:set-up}
\paragraph{Dataset.} We use five datasets, namely Stanford Sentiment Treebank (SST-2~/~SST-5)~\cite{socher2013recursive}, Microsoft Research Paraphrase Corpus (MRPC)~\cite{dolan2004unsupervised}, AG's News Corpus (AG News)~\cite{zhang2015character} and CoNLL 2003 Named Entity Recognition dataset (CoNLL'03)~\cite{sang2003introduction} for experiments.
The statistics can be found in Table~\ref{table:dataset}.
The train/development/test sets follow the original settings in those papers.
We use accuracy for sequence classification and f1-score for sequence labeling as the evaluation metric.

\paragraph{Baseline Approaches.}
We use two common baseline approaches in NLP active learning to compare with our framework, namely random sampling (RM) and entropy-based uncertainty sampling (US).
For sequence classification tasks, we adopt the widely used Max Entropy (ME)~\cite{berger1996maximum} as uncertainty measurement:
\begin{equation}\small
H^{ME}(\mathbf{x})=-\sum_{m=1}^{c}P(\mathbf{y}=m|\mathbf{x})\log P(\mathbf{y}=m|\mathbf{x})
\end{equation}
where $c$ is the number of classes.
For sequence labeling tasks, we use total token entropy (TTE)~\cite{settles2008analysis} as uncertainty measurement:
\begin{equation}\small
H^{TTE}(\mathbf{x}) = -\sum_{i=1}^N\sum_{m=1}^lP(\mathbf{y}_i=m|\mathbf{x})\log P(\mathbf{y}_i=m|\mathbf{x})
\end{equation}
where $N$ is the sequence length and $l$ is the number of labels.

\paragraph{Latent Space Definition}
We use the adversarial attack in our \framework to find informative samples, which rely on a well-defined latent space. Two types of latent spaces are defined here based on the encoder architectures and tasks:
\begin{enumerate}
    \item For pre-trained LMs like BERT~\cite{devlin2018bert}, which has an extra token \texttt{[CLS]} for sequence classification, we directly use its latent state $\mathbf{x}$ as the representation of the whole sentence in the latent space $\mathcal{H}$.
    \item For the other circumstances where no such special token can be used, a mean-pooling operation is applied to the encoder output, i.e. $\mathbf{x} = \frac{1}{n}\sum_{t=1}^nh_t$, where $h_t$ denotes the contextual word representation of the $t_\text{th}$ token produced by the encoder. 
    The latent space $\mathcal{H}$ is spanned by all the latent states.
\end{enumerate}

\begin{table*}[t]\small
\centering
\begin{tabular}{@{}lllllll}
\toprule
& Label Size & 2\% & 4\% & 6\% & 8\% & 10\% \\ \midrule
\multirow{3}{*}{SST-2}&RM& 87.78(.003) & 89.85(.004) & 89.85(.010) & 89.69(.004) & 90.26(.008) \\
&US& 87.74(.004) & \textbf{90.25}(.006) & 90.38(.008) & 90.25(.006) & 91.27(.007) \\
& AUSDS (FGV) & \textbf{89.18}(.002) & 89.88(.008) & 89.16(.014) & \textbf{91.07}(.005) & 89.95(.003) \\ 
& AUSDS (DeepFool) & 88.74(.004) & 90.06(.003) & 89.84(.007) & 90.74(.006) & \textbf{91.58}(.002) \\
& AUSDS (C\&W) & 87.97(.003) & 89.95(.005) & \textbf{90.83}(.007) & 90.12(.003) & 91.13(.001) \\ \midrule
\multirow{3}{*}{SST-5} & RM & 49.45(.010) & 50.01(.007) & 50.88(.006) & 50.39(.014) & 51.35(.005) \\
& US & 49.10(.008) & 49.54(.009) & 50.63(.008) & 50.90(.012) & \textbf{51.43}(.005) \\
& AUSDS (FGV) & 49.57(.006) & 50.36(.008) & 50.09(.009) & 50.19(.014) & 50.62(.011) \\ 
& AUSDS (DeepFool) & \textbf{50.20}(.012) & \textbf{51.87}(.003) & \textbf{51.74}(.012) & 50.97(.012) & 51.23(.007) \\
& AUSDS (C\&W) & 48.28(.012) & 48.78(.014) & 51.58(.007) & \textbf{51.40}(.010) & 47.42(.006)\\ \midrule
\multirow{3}{*}{MRPC} & RM & 67.33(.008) & 68.31(.006) & 68.56(.018) & 70.06(.021) & 71.15(.020) \\
& US & 62.14(.090) & \textbf{69.34}(.005) & 69.11(.010) & 70.53(.017) & 71.49(.016) \\
& AUSDS (FGV) & \textbf{68.89}(.014) & 69.30(.023) & 70.28(.015) & 70.06(.012) & 69.30(.019)\\ 
& AUSDS (DeepFool) & 67.92(.009) & 68.88(.017) & 69.68(.017) & \textbf{71.69}(.014) & \textbf{71.55}(.012) \\
& AUSDS (C\&W) & 67.91(.014) & 68.53(.017) & \textbf{70.46}(.012) & 70.49(.012) & 68.89(.016)\\ \midrule
\multirow{3}{*}{AG News} & RM & 89.89(.003) & 90.89(.002) & 91.37(.002) & 91.79(.002) & 92.21(.002) \\
& US & 90.29(.006) & 91.59(.007) & 92.34(.003) & 92.71(.001) & 93.01(.001) \\
& AUSDS (FGV) & \textbf{90.75}(.002) & 91.55(.002) & 92.26(.003) & 92.62(.001) & \textbf{93.16}(.001) \\ 
& AUSDS (DeepFool) & 90.67(.004) & \textbf{91.65}(.004) & \textbf{92.43}(.004) & 92.66(.004) & 93.12(.002) \\
& AUSDS (C\&W) & 90.24(.002) & 91.29(.002) & 92.30(.004) & \textbf{92.90}(.002) & 93.10(.003)\\ \midrule
\multirow{3}{*}{CoNLL'03} & RM & 80.42(.002) & 83.38(.002) & 85.39(.005) & 86.78(.005) & 87.42(.003) \\
& US & 78.12(.002) & 81.49(.019) & 84.45(.004) & 86.73(.008) & 87.79(.004) \\
& AUSDS (FGV) & \textbf{80.65}(.006) & \textbf{83.60}(.003) & \textbf{85.98}(.010) & \textbf{87.10}(.004) & \textbf{87.83}(.003) \\ 
& AUSDS (DeepFool) & --- & --- & --- & --- & --- \\
& AUSDS (C\&W) & --- & --- & --- & --- & --- \\ \bottomrule
\end{tabular}
\caption{The convergence results w.r.t. the label size in the training from scratch setting with BERT as the encoder. The label size denotes for the ratio of labeled data. The numbers are the averaged results of 5 runs on the test set. The best results with each label size are marked as bold. The sequence classification and sequence labeling tasks are evaluated with accuracy and f1 score, respectively. The AUSDS using DeepFool and C\&W on CoNLL'03 are omitted because these adversarial attack methods are not suitable for sequence labeling.}
\label{table:fromscratch}
\end{table*}

\begin{table*}[!t]\small
\centering
\begin{tabular}{lllllll}
\toprule
& Label Size & 2\% & 4\% & 6\% & 8\% & 10\% \\ \midrule
&RM& 81.58(.004) & 82.90(.006) & 83.53(.008) & 82.15(.016) & 84.40(.006) \\
&US& 78.23(.007) & 80.34(.003) & 81.99(.006) & 82.34(.008) & 82.21(.004) \\
& AUSDS (FGV) & 81.22(.004) & 83.25(.001) & \textbf{84.18}(.005)	& 84.49(.004) & 84.62(.009) \\ 
& AUSDS (DeepFool) & \textbf{82.37}(.003) & 83.31(.004) & 83.77(.002) & \textbf{84.68}(.001) & \textbf{84.73}(.005) \\
& AUSDS (C\&W) & 81.27(.006) & \textbf{84.02}(.007) & 82.76(.002) & 84.40(.002) & 83.58(.012) \\ \bottomrule
\end{tabular}
\caption{The convergence results w.r.t. the label size in the training from scratch setting with ELMo as encoder on SST-2. The label size denotes for the ratio of labeled data. The best results with each label size are marked as bold.}
\label{table:fromscratchelmo}
\end{table*}

\paragraph{Implementation Details.}
We implement our frameworks based on ${\rm BERT}_{\rm BASE}$ model\footnote{https://github.com/huggingface/pytorch-pretrained-BERT} and ${\rm ELMo}$\footnote{https://github.com/allenai/allennlp}.
The configurations of the two models are the same as reported in \cite{devlin2018bert} and \cite{peters2018deep} respectively.
The implementation of the KNN search is introduced in section 3.3.
For the rest hyperparameters in our framework, 1) the batch size and the size of $Q$ is set as 32 (16 on MRPC dataset); ~2) the fine-tuning interval $j$ and the fine-tuning step size $k$ are set as 50 steps; ~3) the ratio $q$ is set as 0.3.
All the tuning experiments are performed on the dev sets of five datasets.
The accumulated labeled data set $D$ is initialized the same for different approaches, taking 0.1\% of the whole unlabeled data (0.5\% for MRPC because the dataset is relatively small).

\begin{figure*}[ht]
    \centering
    \subfigure[Margin during Training]{
        \includegraphics[width=0.86\columnwidth]{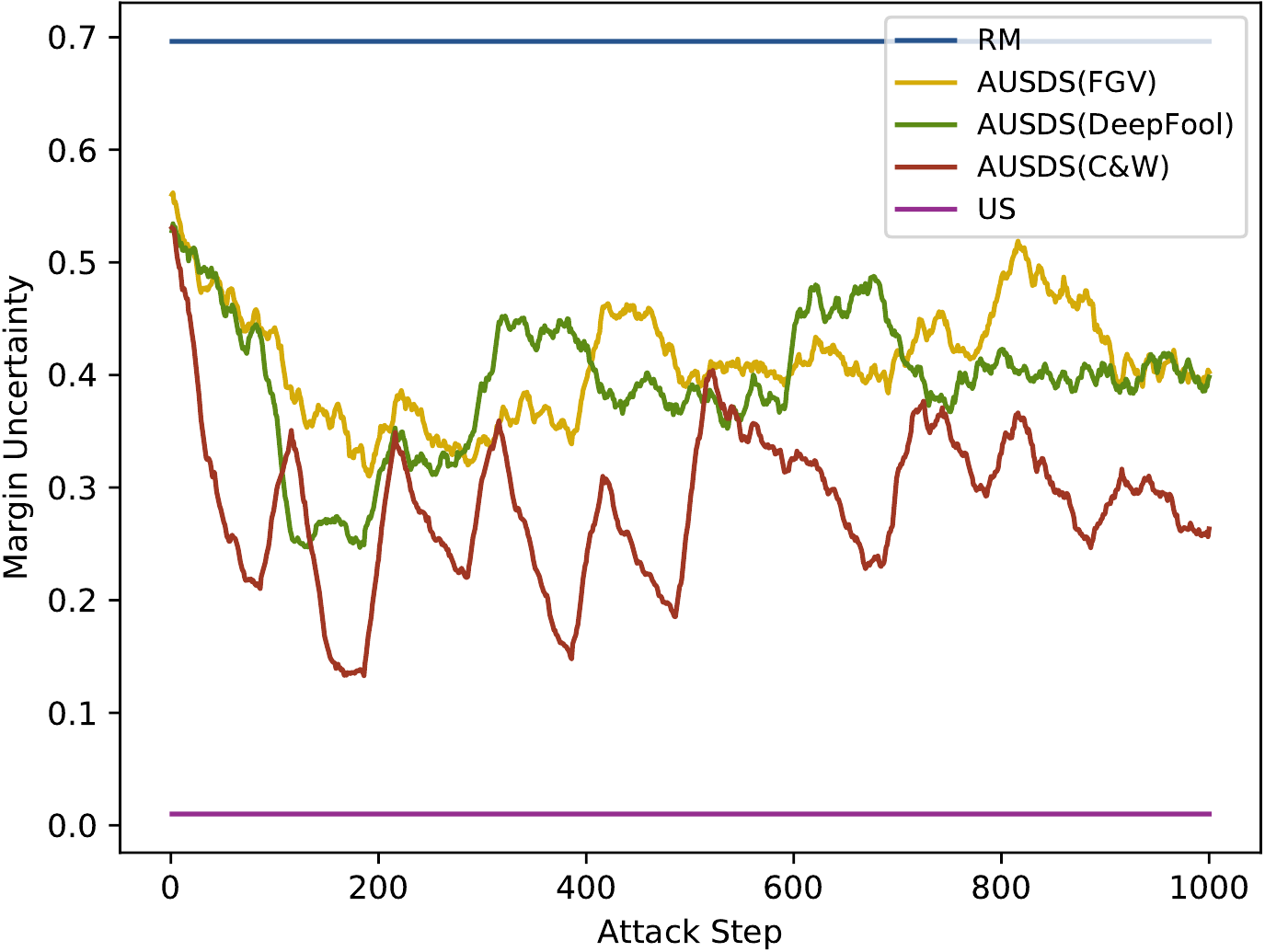}
        \label{fig:curve}
    }
    \subfigure[Margin Distribution]{
        \includegraphics[width=0.86\columnwidth]{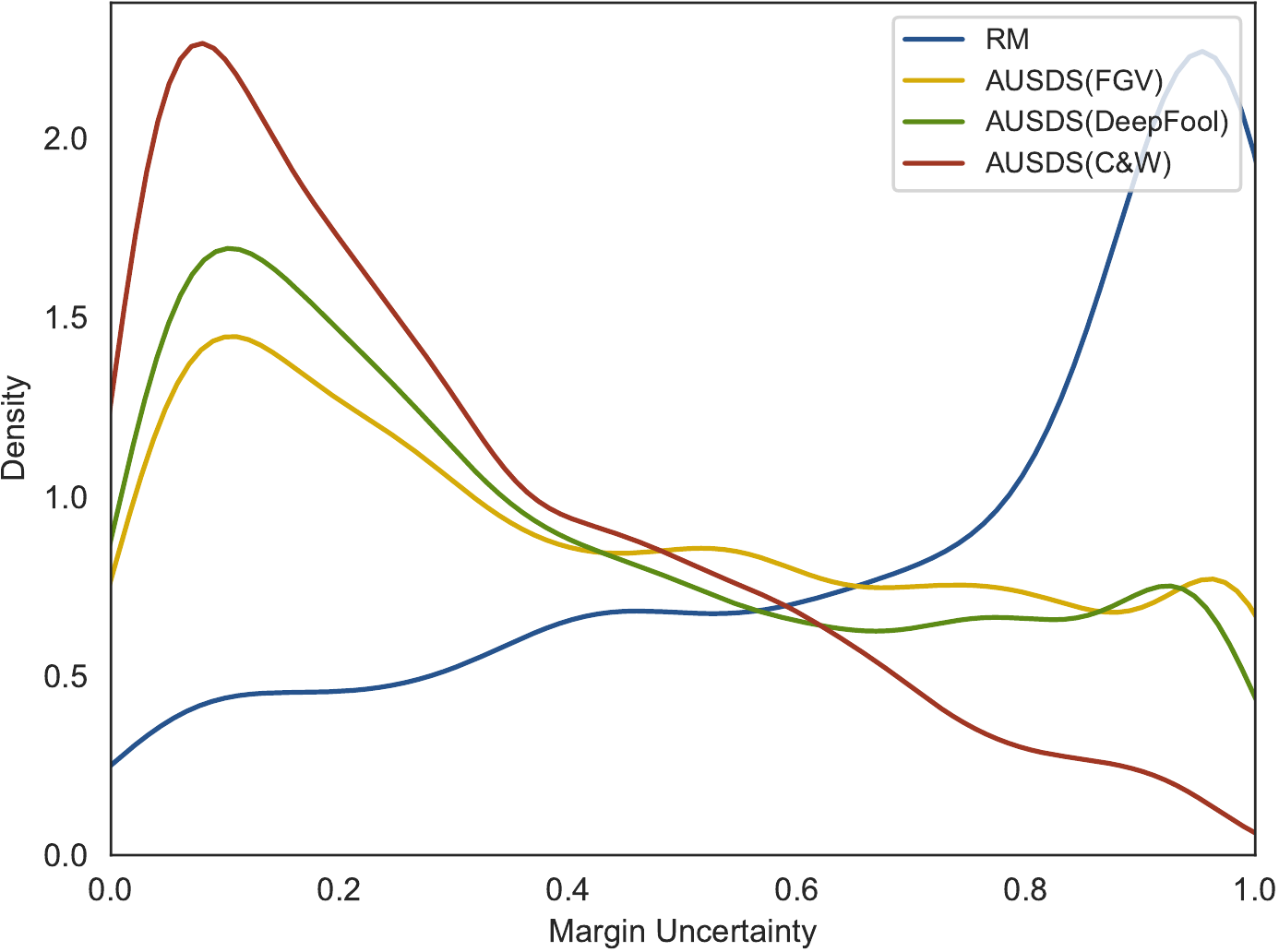}
        \label{fig:dist}
    }
    \caption{The margin of outputs on samples selected by different sampling strategies on SST-5. The margin denotes for differences between the largest and the second-largest output probabilities on different classes. The lower the margin is, the closer the sample is located to the decision boundary. Fig. (a) shows the average margin of each sampling step during training. The margins of samples selected by RM and US on whole unlabeled data are also plotted as references. Fig. (b) shows the margin distribution of samples selected from sampling step 800 to 1000, where the average uncertainty becomes steady. US in Fig. (b) is omitted for better visualization.}
    \label{fig:margin}
\end{figure*}

\subsection{Sampling Effectiveness}
\label{sec:exp:main:effectiveness}
\textbf{\method can achieve higher sampling effectiveness than uncertainty sampling due to the sampling bias problem.}
The main criteria to evaluate an active learning approach is the sampling effectiveness, namely the model performance with a limited amount of unlabeled data being sampled and labeled. Our \framework is compared with the two baselines using the same amount of labeled data. The limitations are set as 2\%, 4\%, 6\%, 8\%, and 10\% of all labeled data in each dataset. We only include at most 10\% of the whole training data labeled, because active learning focuses on training with a quite limited amount of labeled data by selecting more valuable examples to label. It makes no difference whether to perform active learning or not with enough labeled data available. We believe that with less labeled data, the performance gap, namely the difference of sampling effectiveness is more obvious.

We propose training from scratch setting to better evaluate the sampling effectiveness, in which models are trained from scratch using the labeled data sampled by different approaches with various labeled data sizes. We argue that simply training the model until convergence after each sampling step, which we call continuous training setting, can easily induce the problem of sampling bias~\cite{huang2010active}. Biased models in the early training phase lead to worse performance even after more informative samples are given. Thus the performance of models during sampling cannot reflect the real informativeness of selected samples.

The from-scratch training results are shown in Table~\ref{table:fromscratch}.
Our framework outperforms the random baselines consistently because it selects more informative samples for identifying the shape of the decision boundary.
Also, it outperforms the common uncertainty sampling in most cases with the same labeled data size limits because the frequent sampling processes in our approach alleviate the sampling bias issue. Uncertainty sampling suffers the sampling bias problem because of frequent variation of the decision boundary in the early phase of training, which results in ineffective sampling. The decision boundary is merely determined by a small number of labeled examples in the early phase. And the easily biased decision boundary may lead to the sampling of high uncertainty samples given the current model state but not that representative to the whole unlabelled data.
With the overall results on the five standard benchmarks of 2 NLP tasks, we observe that our \method can achieve better sampling effectiveness with DeepFool for sequence classification and FGV for sequence labeling. The results of CW are also included for completeness and comparison.

To prove that our AUSDS framework does not heavily depend on BERT, we conduct experiments on SST-2 with ELMo as the encoder, which has a different network structure. The results in Table ~\ref{table:fromscratchelmo} show that in this setting, our AUSDS framework still achieves higher sampling effectiveness, while the original uncertainty sampling gets stuck in a more severe sampling bias problem. The results in this experiment can also be evidence of the generalization ability of our framework to other pre-trained LM encoding space.

\subsection{Computational Efficiency}

\textbf{\method is computationally more efficient than uncertainty sampling.}
Our \method is computationally efficient enough to be performed at batch-level, thus achieving real-time effective sampling. The average sampling speeds of different approaches are compared w.r.t. US (Table~\ref{tab:samp_speed}).

We observe that uncertainty sampling can hardly work in a real-time sampling setting because of the costly sampling process. Our AUSDS are more than 10x faster than common uncertainty sampling. The larger the unlabeled data pool is, the more significant the acceleration is.
Our framework spends longer computation time, compared with the random sampling baseline, but still fast enough for real-time batch-level sampling. Moreover, the experimental results on Sampling Effectiveness in Sec.~\ref{sec:exp:main:effectiveness} show that the extra computation for adversarial samples is worthy with obvious performance enhancement on the same amount of labeled data.

\subsection{Samples Uncertainty}
\textbf{\method can actually select examples with higher uncertainty.}
We plot the margins of outputs of samples selected with different sampling strategies on SST-5 in Fig.~\ref{fig:margin}. We use margin as the measurement of the distance to the decision boundary. Lower margins indicate positions closer to the decision boundary. As shown in Fig.~\ref{fig:curve}, the samples selected by our AUSDS with different attack approaches achieve lower average margins during sampling. Samples from step 800 to 1000 are collected to estimate the margin distribution, as shown in Fig.~\ref{fig:dist}. It is shown that our AUSDS has better capability to capture the samples with higher uncertainty as their margin distributions are more to the left. The uncertainty sampling performed on the whole unlabeled data gets the most uncertain samples. However, it is very time-consuming and can not be applied frequently.

In short, \method achieves better sampling effectiveness in comparison with US because the more efficient batch-level sampling alleviates the problem of sampling bias. Adversarial attacks can be an effective way to find critical data points near the decision boundary.

\section{Conclusion}
\label{sec:conclusion}
Uncertainty sampling is an effective way of reducing the labeled data size in sentence learning.
But uncertainty sampling of high latency may lead to an ineffective sampling problem.
In this study, we propose adversarial uncertainty sampling in discrete space for active sentence learning to address the ineffective sampling problem.
The proposed \method is more efficient than traditional uncertainty sampling by leveraging adversarial attacks and projecting discrete sentences into pre-trained LM space.
Experimental results on five datasets show that the proposed approach outperforms strong baselines in most cases, and achieve better sampling effectiveness.

\section*{Acknowledgments}

The corresponding author is Yong Yu. 
The SJTU team is supported by "New Generation of AI 2030" Major Project 2018AAA0100900 and NSFC (61702327, 61772333, 61632017, 81771937). 
We thank Rong Ye, Huadong Chen, Xunpeng Huang, and the anonymous reviewers for their insightful and detailed comments.

\bibliography{anthology,emnlp2020}
\bibliographystyle{acl_natbib}

\end{document}